\documentclass[sigconf]{acmart}

\usepackage{booktabs} 

\setcopyright{none}

\usepackage{amsmath,amssymb}
\usepackage[linesnumbered,vlined, ruled]{algorithm2e}
\usepackage[noend]{algpseudocode}
\usepackage{array}
\usepackage{caption}
\usepackage{subcaption}
\renewcommand\footnotetextcopyrightpermission[1]{} 
\begin{document}

\title{SafeRoute:\\ Learning to Navigate Streets Safely in an Urban Environment}

\author{Sharon Levy, Wenhan Xiong, Elizabeth Belding, William Yang Wang}
\affiliation{%
  \institution{UC Santa Barbara}
}
\email{[sharonlevy, xwhan, ebelding, william]@cs.ucsb.edu}

\begin{abstract}
Recent studies show that 85\% of women have changed their traveled route to avoid harassment and assault. Despite this, current mapping tools do not empower users with information to take charge of their personal safety. We propose SafeRoute, a novel solution to the problem of navigating cities and avoiding street harassment and crime. Unlike other street navigation applications, SafeRoute introduces a new type of path generation via deep reinforcement learning. This enables us to successfully optimize for multi-criteria path-finding and incorporate representation learning within our framework. Our agent learns to pick favorable streets to create a safe and short path with a reward function that incorporates safety and efficiency. Given access to recent crime reports in many urban cities, we train our model for experiments in Boston, New York, and San Francisco. We test our model on areas of these cities, specifically the populated downtown regions where tourists and those unfamiliar with the streets walk. We evaluate SafeRoute and successfully improve over state-of-the-art methods by up to 17\% in local average distance from crimes while decreasing path length by up to 7\%.
\end{abstract}

\maketitle

\settopmatter{printacmref=false}
\settopmatter{printacmref=false} 
\renewcommand\footnotetextcopyrightpermission[1]{} 
\pagestyle{plain}

\section{Introduction}\label{sec:intro} Many women take alternative and sometimes longer routes than those that are
recommended by navigation applications, such as Google Maps. This is due to fear of harassment or violence when walking alone or with other women on the streets, especially at night. In a street harassment study by Cornell University and Hollaback from 2014, researchers interviewed 4,872 women in the U.S. and found that 85\% of these women have taken a different route home or to their destination in order to avoid potential harassment or assault, and 67\% changed the time they left an event or location~\cite{hollaback}. While those local to an area might know which streets are safe or risky and can take their own ``safe routes'', others visiting a new city will most likely be unaware of the places they should avoid.
Tourists have to research criminal activity ahead of their planned trip in order to stay informed when walking unfamiliar streets. The creation of a safe routing application is more important than ever. Data collected from 61 metropolitan police agencies showed an 11\% increase in homicides in 2016, the second year in a row with such development~\cite{time_crime}. With a safe routing application, women, tourists, students, and others will have the resource to increase their safety and peace of mind when walking in an urban environment.

In this paper, we focus exclusively on non-vehicular travel (by foot, bicycle, etc.). Walkable cities, such as New York City, Boston, San Francisco, London, etc., typically have multiple potential routes between any two points. We want to compute paths that are both short and risk-free and create a balance between the two objectives. Existing state-of-the-art methods also focus on populated cities but have the disadvantage of using crime density maps to create paths, which smooth over smaller clusters of crimes and therefore ignore smaller crime hotspots.

Recent contributions in deep reinforcement learning show the possibility of short path navigation~\cite{DBLP:journals/corr/abs-1804-00168}. 
However, our model aims to go beyond this and has the complex goal of navigating safely in addition to finding a short path. To provide safe routes, we choose a solution based on deep reinforcement learning, which is a natural choice to many data mining problems that require making incremental decisions: instead of requiring supervised signals at every time step, the policy can be refined based on a single long-term reward signal. This allows us to directly train the agent to co-optimize several goals at once. Other existing approaches in safe path-finding do not use reinforcement learning to generate paths and while there are existing classical reinforcement learning systems in route planning such as \cite{wei2014day}, they do not utilize representation learning and policy gradient techniques. 

We propose SafeRoute, a novel solution to the problem using deep RL. The goal of SafeRoute is to provide users with a safe and short route to their destination. The RL agent learns to choose a safe street at each step that leads from a starting to destination intersection. We test our results against paths found using an existing - non-reinforcement - algorithm (SafePath) that also utilizes crime information when calculating safe routes.

SafeRoute makes three contributions to the problem of safe path-finding:
\begin{description}
  \item[$\bullet$] We are the first to consider deep reinforcement learning to solve the problem of multi-objective path-finding.
  \item[$\bullet$] We propose a reward function for the optimization of safety and length in generated paths. As opposed to using geographical coordinates, we utilize graph embeddings based on local street connectivity to represent maps to create an improved learning environment for our agent.
  \item[$\bullet$] Our model produces results with up to 17\% relative improvement in local average distance from crimes over the state-of-the-art model with samples from maps of Boston, New York, and San Francisco.
\end{description}
  
\section{Related Work}\label{sec:relatedwork}

\noindent
\textbf{Safe Routing:}
There are existing studies that explore the balance between safety and efficiency in routing. One such application, SafePath, creates safe paths for users and optimizes for both safety and distance~\cite{Galbrun:2016:UNB:2894858.2895106}. Using a crime density map, SafePath assigns risks to streets and outputs paths varying from shortest to safest. Another safe path application was developed for Mexico City using tweets and official crime reports to classify and geocode crimes with a naive Bayes classifier~\cite{mexico}. However, it does not consider geographical distance in its algorithm and only focuses on creating paths based on safety. Similarly, SocRoutes creates safe routes using geocoded tweets and routes users through points around unsafe regions~\cite{kim2014socroutes}. SocRoute performs sentiment analysis on tweets and categorizes regions on a map as unsafe when they have more negative tweets on average. To generate its paths, it first creates the shortest path to the destination and, if the path goes through an unsafe region, it moves its waypoints to be outside of that region. While these approaches focus on crimes on a larger scale, our model concentrates on crimes at the street level. A user would most likely not want to deviate from the shortest path by too much, so previous works would either ignore smaller crime hotspots or route users in a much longer distance around a larger crime area.  \\

\noindent
\textbf{Multi-Preference Routing:}
Multi-preference or multi-criteria routing algorithms optimize routes for users with more than one objective, such as our model. One such variety considers both distance and either happiness, beauty, or quietness when generating paths by re-ranking a top-k list of shortest paths according to the second objective~\cite{quercia2014shortest}. However, if the top-k list of paths is not diverse enough from the shortest path, it may be hard to truly optimize the second objective. One of the more common models of multi-preference routing optimizes distance and traffic conditions. For instance, the PreGo system creates paths based on multiple user preferences such as time, scenery, and road conditions through a single traversal of the graph-based map~\cite{hendawi2015multi}. Though this system can route based on risk factor, it does not incorporate distances from risks near edges, which does not give enough information about street safety. Another system, T-Drive, optimizes distance and time using GPS taxi information along with distance to model routes~\cite{Yuan:2010:TDD:1869790.1869807}.  However, there is no explicit way to include crime information in this process. The ARSC algorithm is introduced in~\cite{kriegel2010route} for finding non-dominated paths using a best first graph search. Though these models create paths based on multiple criteria, most do not utilize crime information while routing, which can lead to the generation of unsafe paths. In addition, our model's continuous state space allows us to incorporate various attributes into our state and integrate representation learning of the network with explicit path-finding. Multi-criteria optimization falls into the category of NP-Hard problems and is not solvable by theoretical computer science. However, the usage of deep reinforcement learning allows us to perform approximation within the learning to search framework, acting as a more practical solution. Another reason for the use of deep reinforcement learning is the ability to integrate human feedback into the model. As risk is not deterministic, other non-deep reinforcement learning approaches fail as they need explicit functions to define risk. Human feedback can be used to provide policy updates and to continuously learn and shape the model even more towards a human-generated path. To elaborate, users can decide whether or not they like a provided path and based on this information, the model can be rewarded or punished for future generation of paths in the area. With such advantages, it is clear that our task can benefit from the use of deep reinforcement learning.\\

\noindent
\textbf{Deep RL in Navigation:}
In ~\cite{brunner2017teaching}, a deep reinforcement learning model was developed to localize itself on a 2D map from a 3D perspective and find shortest paths out of their respective mazes. The RPA model uses a combination of model-based and model-free learning for visual navigation with textual instructions~\cite{DBLP:journals/corr/abs-1803-07729}. Recently, Google's DeepMind has created a deep reinforcement learning model that trains on Google Street View in order to navigate cities without a map~\cite{DBLP:journals/corr/abs-1804-00168}. For tasks of reaching a destination point, the model represents the target in relation to its distances from landmarks nearby. One of the drawbacks of image-based navigation is the amount of data required for training. Furthermore, SafeRoute attempts to co-optimize the two goals of safety and distance, which would result in additional training to not only be able to find a target with unstructured image data but also classify unsafe streets and avoid them. Graph-based navigation appears in some deep reinforcement learning frameworks. DeepPath~\cite{wenhan_emnlp2017} uses deep reinforcement learning to infer missing links within a knowledge graph. Another method trains on recordings of maze navigation to build a topological map and later navigate to a destination within the maze~\cite{savinov2018semiparametric}. 

In spite of prior work in safe path-finding and multi-preference routing, SafeRoute and its underlying architecture differentiates itself by 1) analyzing crimes in the direct path area in order to generate its routes, 2) utilizing continuous state space to allow different types of information as input and incorporate representation learning, and 3) enabling human feedback through policy updates.

\section{SafeRoute}\label{sec:model}
We can view the route finding process as a Markov Decision Process. 
At each time step, the agent makes a decision on which compass direction to go next, eventually leading to the final destination. Start and end coordinates of intersections are input into the model, which returns a list of coordinates, relating to the incremental decisions made by the agent. By rewarding the agent for avoiding crime-filled streets, we create a safe path for the user. The following section describes the framework and the training and testing pipeline behind the deep reinforcement learning architecture of SafeRoute. The environment and policy-based agent are discussed, along with the rewards system and used to find short and safe paths within a map. In addition, the two forms of training (supervised and retraining with rewards), along with the testing algorithm are described.

{\setlength\intextsep{0pt}
\begin{figure}[!t]
\includegraphics[height=2.1in]{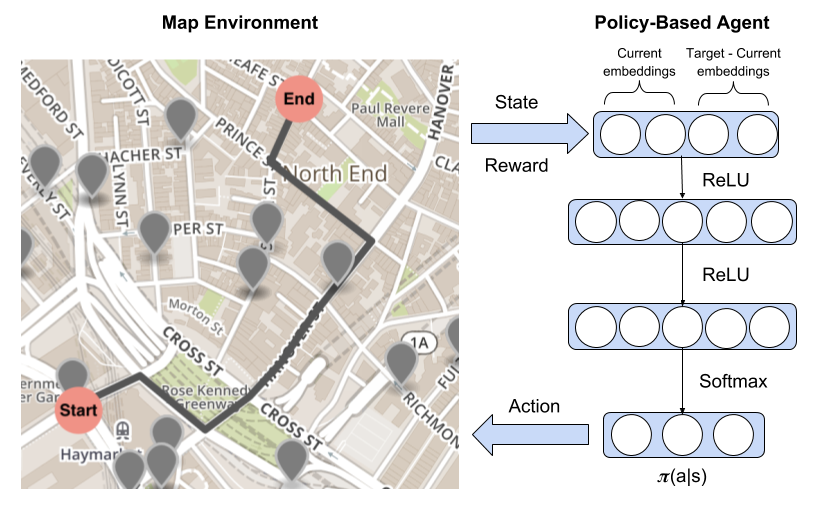}
\caption{The SafeRoute model containing the map environment and policy-based agent. The current state is fed into the agent's neural network and the action to take is outputted. Once the action is taken in the environment, the reward is collected and used to update the agent's policy.}
\label{fig:architecture}
\end{figure}
}%

\subsection{Architecture}
The SafeRoute system is split into two parts: an environment with which the RL agent interacts, and the policy network the RL agent represents and uses to make decisions within the environment. The SafeRoute architecture can be seen in Figure~\ref{fig:architecture}. The environment is represented as a Markov Decision Process with tuple $<S, A, P, R>$. $S$ represents the continuous states of the environment and $A = \{{a_1, a_2, ...a_N}\}$ defines all actions available to the agent. $P(S_{t+1} = s_0|S_t = s, A_t = a)$ determines the probability of moving from one state to another. $R(s,a)$ is the function that rewards the agent for taking action $a$ when in state $s$. \\

\noindent
\textbf{States: }
The agent's state represents its current status on the map. In our model, the state uses the agent's current and target position. Including the target position allows the agent to relate actions to the end goal so if it is in the current position at a later point with a destination in the opposite direction from before, the agent will not take the same actions. This also allows the agent to generalize better to new unseen states. If the new target is near a target the agent has been trained on, the agent will take similar actions towards the new goal when starting from the same area. In order to represent states, map information is transformed into a graph with street intersections as nodes and streets connecting two nodes as edges. The graph is directed and edges have a compass direction. Graph embeddings are used to represent the continuous states of the RL agent instead of the latitude/longitude coordinates. These embeddings are generated using node2vec~\cite{grover2016node2vec} on the
graph. The reasoning behind using graph embeddings as opposed to coordinates is that coordinates do not give any information as to how intersections are connected to each other on the actual map. In the initial iterations of the project, coordinates were used to represent states. However, even with supervised training, our model was unable to learn to navigate the map. With the graph embeddings, the model is able to determine the streets that lead to certain intersections and eventually, the final destination.
The states use embeddings from the agent's current node and target node as follows: 
$$\textbf{s}_t = (\textbf{e}_t, \textbf{e}_{target} - \textbf{e}_t)$$
where $\textbf{e}_t$ denotes the embeddings of the current node and  $\textbf{e}_{target}$ denotes the embeddings of
the target node. \\

\noindent
\textbf{Actions: }
Actions in the environment are represented as moving from one street intersection to another. The actions themselves are compass directions (North, Northeast, East, Southeast, South, Southwest, West, Northwest). The RL agent learns to pick actions, out of all available actions, that lead in the direction of the target intersection while also moving away from high crime areas. \\

\noindent
\textbf{Policy Network: }
The policy network representing the RL agent uses a stochastic policy, $\pi_\theta(s, a) = p(a|s; \theta)$, where $\theta$ is the list of neural network parameters that are updated using the Adam Optimizer~\cite{kingma2014adam}. A stochastic policy is used instead of a greedy policy to prevent the agent from getting stuck in cycles on the map. Using a stochastic policy, the agent can break free of cycles such as repeatedly moving towards a dead end that appears to be leading in the right direction or continually taking a path that will eventually result in a dead end. The neural network contains two hidden layers, each with a rectified linear unit (ReLU) activation function. The output uses a softmax function and returns a probability distribution over all actions. The actions are pruned for those that are not available at the current intersection, and the remaining probabilities are normalized and returned. At a high level, the neural network takes as input a state $s$ and outputs a normalized probability distribution over all available actions. \\

\noindent
\textbf{Rewards: }
The agent optimizes for multiple preferences, so the reward function must consider a number of different factors. Since an integral feature of SafeRoute is to avoid crime areas when creating a route, we add safety into the reward as a function of average distances from previously known crime scenes.  A list of recent crimes in the city is traversed for those at a certain radius from the current location. The radius and location vary per edge along the path. The location used is the midpoint along each edge and the radius is equal to the length of that edge. The average distance from crimes within the radius at each step is then calculated. The average distance is used as opposed to directly calculating the number of crimes because we value more crimes at the edge of the radius as better than fewer crimes directly along the route. 

Although the primary goal of SafeRoute is to increase safety by generating routes that lead away from high crime areas, we also want to consider efficiency in our reward represented by the path length. The total average distance from crimes along the path is calculated and then divided by the path length. If there are no crimes near the path, $\kappa$ 
is used as a reward. All other paths are assigned a reward proportional to their average distance from crimes divided by the path length. The final reward is defined below:

\[
	r_{CRIME} = 
\begin{cases}
  \frac{\frac{\sum_{i = 1}^{n}\sum_{j = 1}^{m}distance(x_i,c_{ij})}{\sum_{i = 1}^{n} number(c_i)}}{length(p)},& \text{if } c \neq \emptyset\\
    \kappa,              & \text{otherwise}
\end{cases}
\]
\noindent
where $n$ is the number of edges along the path, $m$ is the number of crimes within the radius at each node, $x$ is the list of edge midpoints along the path, $c$ is the list of crimes in each radius, $p$ is the path, and $\kappa$ is a hyperparameter. With this reward in place, shorter paths will be rewarded more than longer paths with similar crime rates. 

\subsection{Training}
Training for SafeRoute comes in two parts: supervised training and retraining with rewards. Without supervised training as the initial step, the agent will have a hard time finding a path to the target node and can end up wandering in random directions. AlphaGo~\cite{Silver_2016} uses imitation learning~\cite{Hussein:2017:ILS:3071073.3054912} as the first step in its training process in order to give the agent an initial push when starting to train with rewards. Similarly, we also start the training pipeline of SafeRoute with supervised training as its form of imitation learning. \\

\noindent
\textbf{Supervised Training}
In SafeRoute, one of the criteria for a good path is short distance. Therefore, we use supervised training to feed in shortest paths for each training episode. The training samples used include a randomly sampled starting intersection on the map and several endpoints at a 5-hop distance on the respective graph. These samples are shuffled and each episode uses Dijkstra's algorithm~\cite{dijkstra1959note} for shortest paths using the edge length as the weight. At the end of each episode, the neural network parameters $\theta$ are updated to reward the actions taken at each state. Each state-action pair within the path is rewarded equally using Monte-Carlo Policy Gradient (REINFORCE)~\cite{Williams1992}.
The reward given in each episode is 1 so our final gradient when updating the policy is equal to:
\begin{align*}
\triangledown_{\theta} J(\theta) &= \sum_t\sum_{a \epsilon A} \pi(a|s_t;\theta) \triangledown_{\theta} log  \pi (a|s_t; \theta) \\
&\approx \triangledown_{\theta} \sum_t log \pi (a = a_t | s_t; \theta)
\end{align*}

\noindent
where $a_t$ is the corresponding action taken at time $t$ along the path. \\

\SetKw{KwBy}{by}
\begin{algorithm}[t]
    \For{$episode\gets1$ \KwTo $N$}{
  		Initialize $max\_rwd\gets0.0$ \\
        Initialize $avg\_rwd\gets0.0$ \\
        Initialize $num\_success\gets0.0$ \\
        Initialize $max\_path\gets\emptyset$ \\
        \For{$i\gets0$ \KwTo $T$} {
        	Initialize state vector $s_t\gets{s_0}$ \\
            Initialize episode length $num\_steps\gets0$ \\
            \While{$num\_steps < max\_len$} {
            	Randomly sample action $a \sim \pi(a|s_t)$ \\
                Add $<s_t, a>$ to $path$ \\
                \If{$success$ or $num\_steps = max\_len$} {
                	\If{$success \text{ and } R_{CRIME} > max\_rwd$} {
                    	$max\_rwd = R_{CRIME}$ \\
                        $max\_path = path$
                    }
                    break \\
                }
                Increment $num\_steps$ \\
                $s_t\gets{s_{t+1}}$ \\
            }
        }
        \If{$max\_rwd \neq 0.0$} {
        	$b\gets{\dfrac{avg\_rwd}{num\_success \text{ or } 1}}$ \\
        	\For{$<s_t, a>$ in $path$} {
            	 Update $\theta$ with $g\propto$ \\
                 $\triangledown_{\theta} \sum_t log \pi (a=r_t | s_t; \theta) (R_{p-(t-1)} - b)$ \\
            }
            Increment $num\_success$ \\
            $avg\_rwd\gets{avg\_rwd + max\_rwd}$ \\
        }
    }
    \caption{Algorithm for retraining with rewards}
    \label{alg:train}
\end{algorithm}

\noindent
\textbf{Retraining with Rewards}

\noindent
After supervised training, the agent is retrained to avoid crime filled areas. We run the model $T$ times for every episode. Due to the stochastic nature of the policy, all $T$ paths found by the agent will likely exhibit some variation. This allows us to have a variety of paths to choose from to reward. For each of the successful paths generated, we look at the rewards calculated along the path. When updating the policy we only consider the path with the highest rewards. No updates are made for the other successful and non-successful paths. Algorithm~\ref{alg:train} illustrates the training procedure.

Instead of immediately updating the policy with the rewards for the best path, we use a baseline value $b$ such as the one used in ~\cite{DBLP:journals/corr/ZarembaS15}. The baseline we use is a running average of the current rewards in the current epoch, using the rewards from the most successful path in each episode. With the baseline in place, the most successful path from each episode will be rewarded only if its rewards are greater than the baseline.

In addition to using a baseline value, we also do not reward each state-action pair equally. Instead, each is rewarded for its actual value in creating the path. At each time step $t$, the current action is rewarded with only the remaining path in mind. Any information coming from previous edges along the path is not included in the calculation of the reward.  The final gradient is shown below:
$$\triangledown_{\theta} J(\theta) =  \triangledown_{\theta} \sum_t log \pi (a = r_t | s_t; \theta) (R_{p-(t-1)} - b)$$
\newline

\SetKw{KwBy}{by}
\begin{algorithm}
    \For{$episode\gets1$ \KwTo $N$}{
    		Initialize state vector $s_t\gets{s_0}$ \\
            Initialize environment $e_t\gets{e_0}$ \\
            Initialize probability $p_t\gets{1.0}$ \\
            Initialize current paths $curr\_paths\gets{(s_t, e_t, p_t)}$ \\
            Initialize successful paths $success\_paths\gets{\emptyset}$ \\
            Initialize episode length $num\_steps\gets0$ \\
            Initialize new paths $new\_paths\gets{\emptyset}$ \\
            \While{$num\_steps < max\_len$ and $|success\_paths| \neq 5$} {
            	$new\_paths\gets{\emptyset}$ \\
            	\For{$s_t, e_t, p_t$ in $curr\_paths$} {
                	\For{$i\gets1$ \KwTo $5$} {
                		Randomly sample action $a \sim \pi(a|s_t)$ with probability $p$ \\
                    	Copy $s_t, e_t$ to $s_c, e_c$ and take action $a$ in new environment \\
                        $p_c\gets{p_t * p}$ \\
                        \eIf{$success$} {
                        	Add $s_c, e_c, p_c$ to $success\_paths$ \\
                        } {
                        Add $s_c, e_c, p_c$ to $new\_paths$ \\
                        $s_t\gets{s_{t+1}}$ \\
                        }
                	}
                }
                \While{$|new\_paths| > 5 - |successful\_paths|$} {
                	Remove $<s_t, e_t, p_c>$ from $new\_paths$ with lowest $p_c$
                }
                Increment $num\_steps$ \\
                $curr\_paths\gets{new\_paths}$ \\
            }
            \If{$|successful\_paths| > 0$} {
        		Choose $path$ from $successful\_paths$ with greatest local average distance from crimes \\ 
                Post-process and return $path$ \\
            }
    }
    \caption{SafeRoute algorithm for testing}
    \label{alg:test}
\end{algorithm}

\noindent
\textbf{Testing}

\noindent
For all the evaluations, we use beam search to create our paths with a beam size of $Z$. The policy runs several times for every current path in the beam and these paths are extended into several new paths. Paths that remain in the beam are those that are highly favored by the policy. Once 5 paths have successfully reached the target or the step limit is reached, the path with the highest local average distance from crimes, as described in Section~\ref{sec:experiments}, is chosen as the final path. Due to the stochastic nature of the agent, the path created will occasionally contain loops that lead back to an intermediate node along the path. Post-processing is done to remove these external loops and the resulting path is returned. Algorithm~\ref{alg:test} illustrates the training procedure.

\section{Experiments}\label{sec:experiments}
\subsection{Dataset}
Though our baseline utilized public crime information for its safe routing, we did not have access to SafePath's dataset and thus created our own for the purposes of our experiments\footnote{Source code and dataset will be available:  https://github.com/sharonlevy/SafeRoute}. Map information was collected from OpenStreetMap, a free collaborative world map~\cite{OpenStreetMap}. We chose to export map information for the downtown areas of Boston, New York, and San Francisco. This is due to the fact that many tourists visit these areas and, as they are urban centers, there are typically multiple viable paths between any two points.  The resulting graphs for Boston and San Francisco were similar in size and were trained for 2,000 episodes per epoch while New York's was larger and trained for 4,000 episodes per epoch. All three models trained for 60 epochs. Hyperparameter tuning was used to determine the value of $\kappa$ in the reward function. We chose five as our value for $T$ in the training algorithm in order to balance the speed of retraining versus the diversity of the generated paths. A beam size of five was used in the testing algorithm. Crime data is collected from Spotcrime, which shows recent crime incident information with details such as geographic coordinates and type of crime~\cite{spotcrime}. For SafeRoute, we chose to use the crimes of shooting, assault, and robbery. 

A study done by ~\cite{yang2012walking} found that walking trips in the U.S. are on average 0.7 miles and with a median of 0.5 miles. Therefore, our model is tested on paths near these lengths. The model is trained on start and end points that are at a 5-hop distance away from each other in the graph. These points represent initial and final points that can be reached in five decisions, though the actual length may vary depending on the city grid and the path chosen. When doing experiments, we randomly sample nodes in the graph and test on 5-hop and 10-hop paths. We chose to test on these number of hops because they resulted in paths ranging from 0.20 - 1.0 miles on average when tested using Dijkstra's shortest path algorithm. Therefore, we cover the spectrum of most walking distances. However, because we are prioritizing both distance and safety, our generated paths are usually longer than their shortest path distance.

\begin{table*}[t]
\centering
\begin{tabular}{{l}{l}|*{3}{c}|*{3}{c}}
\toprule
& & \multicolumn{3}{c}{5-hops} & \multicolumn{3}{c}{10-hops} \\
\hline
City & {Model} & {Local}  & Global  & Length & {Local}  & Global  &Length  \\
\hline
Boston &  Dijkstra & 0.0554  & 0.8620 & 0.2023 & 0.0484 & 0.8479 & 0.4200 \\
 & SafePath(Median)  & 0.0566  & 0.8648 & 0.2044 & 0.0498  & 0.8639  & 0.4372 \\
& SafePath(Safest)  & 0.0568  & \textbf{0.8651}  & 0.2050 & 0.04817 & \textbf{0.8768}  & 0.4619 \\
& SafeRoute   & \textbf{0.0630}  & 0.8627  & 0.2361 & \textbf{0.0543} & 0.8583 &0.5031 \\
\hline
San Francisco &  Dijkstra & 0.0882  & 0.9268 & 0.4704 & 0.09638  & 0.8966 & 1.0121 \\
 & SafePath(Median)  & 0.0934  & 0.9423  & 0.4821 & 0.1046  & 0.9344  & 1.1063 \\
& SafePath(Safest)  & 0.0937  & 0.9489  & 0.5001 & \textbf{0.1084}  & \textbf{0.9650}  & 1.2402 \\
& SafeRoute   & \textbf{0.0990}  & \textbf{0.9880}  & 0.5468 &  0.1036  & 0.9611 & 1.1659 \\
\hline
New York & Dijkstra & 0.1344  & 1.1978 & 0.4222 & 0.1016  & 1.1842 & 0.7907 \\
& SafePath(Median)  & 0.1341  & 1.2125  & 0.4515 & 0.0987  & 1.2267  & 0.8748 \\
& SafePath(Safest)  & 0.1344  & \textbf{1.2242}  & 0.4935 & 0.0978  & \textbf{1.2537}  & 0.9731 \\
& SafeRoute  & \textbf{0.1454}  & 1.2004 & 0.47267 & \textbf{0.1179}  & 1.1992  & 0.8915\\
\bottomrule
 \end{tabular}
 \caption{Results for the average distance from crimes (local), average distance from crimes (global), and length experiments on 5-hop and 10-hop test datasets. Distance is calculated in terms of miles. In our evaluation, larger values for the local and global metrics are better while smaller values for the length are favorable. }\label{tab:results}
\end{table*}

\begin{table*}[t]
\centering
\begin{tabular}{p{2cm}|*{3}{c}|*{3}{c}}
\toprule
&  \multicolumn{3}{c}{5-hops} & \multicolumn{3}{c}{10-hops} \\
\hline
City & {Local(\%)} & Global(\%)  & Length(\%)  & {Local(\%)} & Global(\%)  & Length(\%)  \\
\hline
Boston  & 10.4 & -0.3  & -20.9 & 15.8 & -2.6 & -13.0\\
\hline
San Francisco  & 4.5  & 4.1  & -7.6 & -6.6  & 0.3  & 5.3\\
\hline
New York  & 6.9  & -2.1  & 2.2 & 17.1  & -4.7 & 6.5\\
\hline
\hline
Average  & 7.3  & 0.6  & -8.8 & 8.8  & -2.3  & -0.4\\ 
\bottomrule
 \end{tabular}
\caption{Percent improvement in results over the state-of-the-art model (SafePath) in safest mode. Models are evaluated on local average, global average, and path length metrics. Results used the percent averages of 3 runs.}\label{tab:percent}
\end{table*}

\subsection{Evaluation Settings}

We evaluate our model in a number of experiments. The quality of the paths is measured in three different tests: average distance from crimes (local), average distance from crimes (global), and path length. These are all measured in terms of miles. \\

\noindent
\textbf{Local Average} The local average distance from crimes only considers crimes near the agent as it traverses the path and is calculated similarly to the agent's rewards:
\[
	Avg Crime(local) =
\begin{cases}
  \frac{\sum_{i = 1}^{n}\sum_{j = 1}^{m}distance(x_i,c_{ij})}{\sum_{i = 1}^{n} number(c_i)},& \text{if } c \neq \emptyset\\
    AvgMinCrime,              & \text{otherwise}
\end{cases}
\]
\noindent
where $n$ is the number of edges along the path, $m$ is the number of crimes within the radius at each edge, $x$ is the list of edge midpoints along the path, $c$ is the list of crimes in each radius. If a path happens to have no crimes within its radius, then it uses the average minimum distance from crimes and is consistent with our safety first approach - highly valuing edges with no crimes around them. \\

\noindent
\textbf{Global Average}
The global average distance from crimes experiment is equivalent to the local test but considers all crimes at each edge along the path. This is seen below:
$$ AvgCrime(global) = \frac{\sum_{i = 1}^{n}\sum_{j = 1}^{m}distance(x_i,c_{j})}{n}$$

\noindent
where $m$ is the list of all crimes in the city and $c$ is the list of crimes in the city. \\

\noindent
\textbf{Path Length}
We sum up the edge lengths along a path in order to determine the final path length:
$$ Length = \sum_{i = 1}^{n}length(n) $$

We evaluate SafeRoute against the baselines on the different metrics. We selected these evaluations based on what we believe to be easily explainable criteria that are likely to be valued by humans trying to safely navigate a new city. Though we propose the metrics of both local average and global average distance from crimes, we value the local average distance metric the most. This metric calculates the average distance to a set of crimes, constrained to the crimes that are within a certain distance from the edges traversed by the path. It is very similar to the reward function used by the RL agent and therefore provides a measure of how well the agent learned the given task.
Furthermore, we believe this metric to be a good approximation of the intuitive overall safety of a path, which is why we developed our model to train on it. By keeping a high average distance from local crimes we can measure how well the agent learns to avoid certain unsafe streets and even navigate within the more crime-filled areas of the city. As an additional metric, we also examine the global average distance along our paths as it considers the entire crime list in its score as opposed to the local average distance. However, this metric is less significant when traversing a path. The global average distance captures information from the whole city and users would likely not care about crime hotspots miles away from their paths. 

We compare our model against two baselines: SafePath and Dijkstra's algorithm. SafePath creates safe paths for users by producing non-dominated paths when considering both safety and distance. When generating safe paths, it assigns streets a risk factor determined by a crime density map of the city. This approach outputs multiple paths on a varying scale of safety and distance using Dijkstra's algorithm. SafePath represents paths on a graph by visualizing risk versus length. In order to generate the paths in between, the algorithm re-weights streets using the gradient of the line between two paths. The final paths appear as the lower convex hull on the risk vs. length graph. For our evaluation, we compare against SafePath's safest paths and the median paths that it outputs, meaning that these paths balance safety and distance. We use Dijkstra's algorithm with distance as the edge weights as our comparison against shortest paths, as is done in SafePath.

\subsection{Results}

Table~\ref{tab:results} shows a numerical comparison between the experimental results of SafeRoute and the baselines. It is evident by how different the distances are that each city has a different structure and distribution of crimes. However, even with these variations, SafeRoute is able to perform well in each city with the same rewards function. Comparing the local average crime distances between the three cities for 5-hop paths shows the numbers are much lower for Boston, implying that it is harder to navigate away from crime in this city. When analyzing the crime density per city, this was shown to be true as Boston had the highest crime density with San Francisco following closely behind. Meanwhile, New York had a density of about half of Boston. The crime distance metrics in Table~\ref{tab:results} follow the same pattern with Boston having the shortest distances from crimes and New York with the highest. However, even with this entailment, SafeRoute is able to navigate further from the crime spots than the baselines. When traversing longer distances at a 10-hop radius, it is noticeable that the local average crime distance decreases by about the same factor for Boston and San Francisco, and by a larger one for New York. This implies that traveling longer distances in New York City requires one to go closer to crime spots so that length is not compromised significantly. Nonetheless, SafeRoute finds paths that are further away from local crimes on average for this city.

In order to decrease variance in our results, we create three models for each city and average the results for each. Table~\ref{tab:percent} shows SafeRoute's percent increase/decrease when compared to SafePath's safest routes. Because our agent receives rewards based on its local average distance from crimes, it is no surprise that our model surpasses SafePath by a large percentage. It is also worth noting that while our model increases path lengths for most experiments, as is expected, there were some test sets in which our average length was shorter than the SafePath while still excelling in some experiments. When comparing results for 5-hop and 10-hop paths, it is evident that SafeRoute has better results for the shorter paths due to the smaller search space and the lower likelihood that the agent will take a wrong action before it reaches the goal.

It is expected that we do not have shorter paths than Dijkstra since we are co-optimizing two criterion. Nevertheless, we demonstrate that it is necessary to compromise length to ensure safety along a path, as confirmed in Table~\ref{tab:percent}. The overall results of the experiments imply that our SafeRoute model has learned to successfully avoid dense crime areas on a map by creating paths that are further from local crimes on average and is able to find a balance between distance and safety.

\begin{figure}[t]
\centering
\begin{subfigure}{.5\columnwidth}
  \centering
  \includegraphics[height=1.8in]{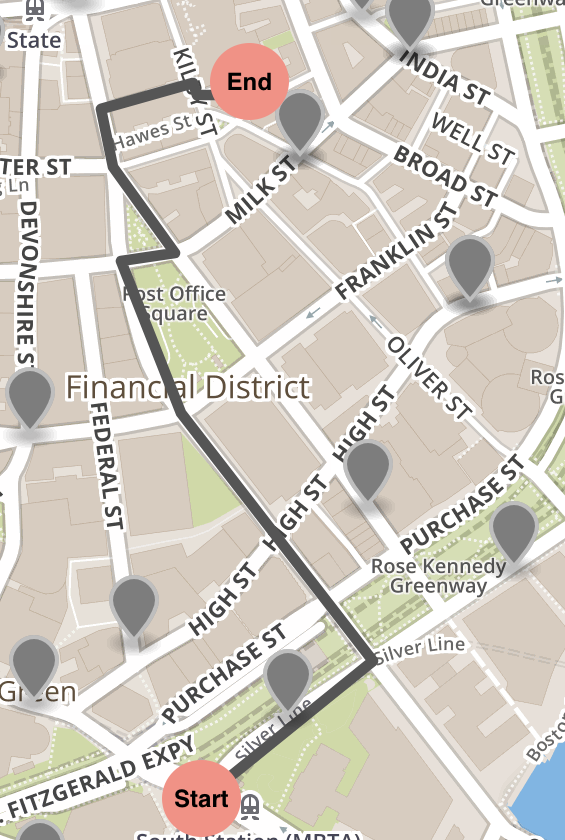}
  \caption{SafeRoute}
  \label{fig:sub1}
\end{subfigure}%
\begin{subfigure}{.5\columnwidth}
  \centering
  \includegraphics[height=1.8in]{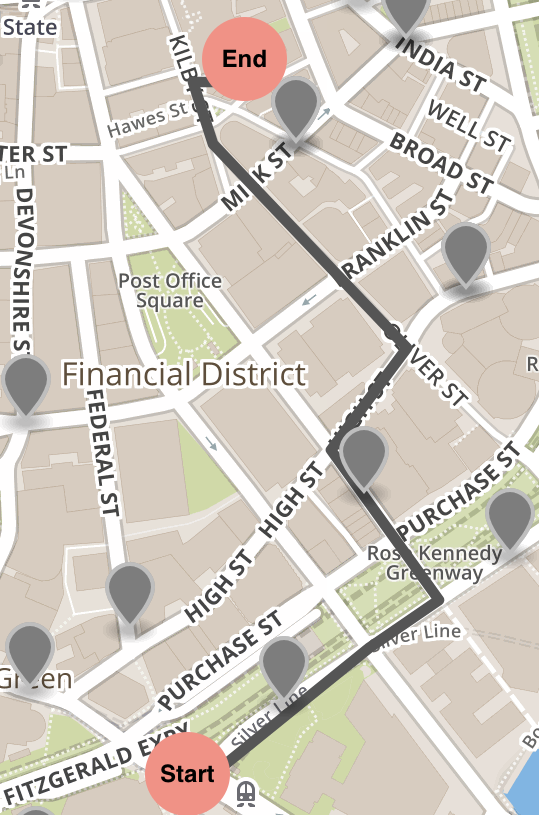}
  \caption{SafePath}
  \label{fig:sub2}
\end{subfigure}
\caption{An example of paths generated by SafeRoute and the state-of-the-art (SafePath) in its safest mode.}
\label{fig:test}
\end{figure}

We also show a sample result from SafeRoute and our baseline on a map of Boston in Figure~\ref{fig:test}. While initially, the two go along the same path, SafeRoute stays in a mostly crime free area for the rest of the path and therefore maps the user away from crimes. Meanwhile, SafePath routes in another direction and ends up closer to crime points instead. 

\subsection{Discussion}
As can be seen in Table~\ref{tab:results}, SafePath evaluation results do not change significantly when attempting to create a safe route vs. a short route. SafePath creates a smooth crime density map in order to utilize Dijkstra's algorithm. The crime density map is created using a Gaussian kernel density estimation (KDE)~\cite{rosenblatt1956}. The density map used by SafePath for our San Francisco dataset is shown in Figure~\ref{fig:density}. It can be seen that the density map has one major peak of crime intensity and does not capture the local variations in crime on the map. Due to the smoothing of the density map, small clusters of crimes go seemingly unnoticed and are visualized as similar to single points of crime as seen in Figure~\ref{fig:density} at (-0.035, 0.014). In addition, most edges in the same area will appear to have very similar risk weights. These weights only have significant change when traversing long distances. As a result, when using Dijkstra's to create risk-free paths for a short distance, as to SafePath, the output will most likely be the one with the fewest hops and longer paths will be overlooked by this algorithm. It is also apparent that the differences between SafePath's safest and median modes are very small. When comparing the results for the local average experiment in Table~\ref{tab:results}, the resulting values are very close. Meanwhile, SafeRoute is shown to have a big increase over the two in its local average metric. This reveals that SafePath's algorithm does not do very well in compromising its length for a safer path. In contrast, SafeRoute is inherently local and puts a high value on avoiding local crime spots. This is done using a non-linear reward system. By rewarding our agent based on local crimes we train it to move towards safer streets that are nearby. With this reward system in place, it is evident that SafeRoute will generate paths that are co-optimized for safety and distance. Because SafeRoute co-optimizes for both parameters, some of our experiments revealed paths that were not longer than those chosen by SafePath in its safest mode. 

{\setlength\intextsep{0pt}
\begin{figure}[t]
\includegraphics[height=2.3in]{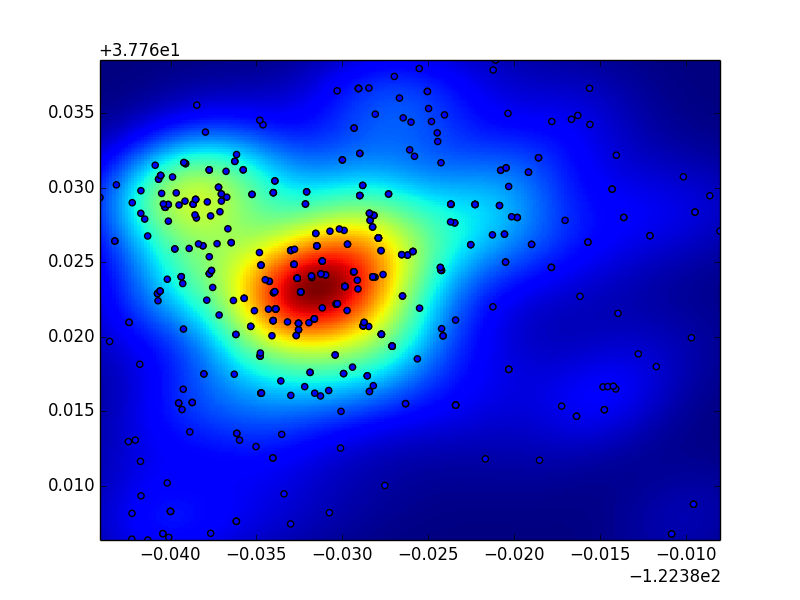}
\caption{Crime density map of San Francisco with the black dots representing crimes. The x-axis and y-axis represent the longitude and latitude coordinates, respectively.}
\label{fig:density}
\end{figure}
}%

As mentioned in Section~\ref{sec:model}, some paths generated by SafeRoute will contain cycles. These decisions will take the agent on a wrong turn, but eventually lead back to an intermediate node along the path and continue to the destination. To handle this, we post-process our paths to remove external loops and return the resulting path. We do not allow the agent to take the same action in a specific state twice to prevent future cycles. Without this, the agent would end up in an endless cycle between two nodes. An example of a looping path before post-processing can be seen in Figure~\ref{fig:error}. When analyzing the path, we see the agent makes a wrong turn on its first step and goes North instead of West towards the destination. This occurs again shortly after. However, after the initial steps, the agent is able to directly reach the goal in an optimal path. To better understand this, we examined each step along the path and found that the agent had high probabilities of going North initially, which decreased as the steps progressed towards the goal and switched to maximizing probabilities for the correct actions. One explanation for this is that not enough training samples crossed the initial area of the path. As a result, the agent maximized probabilities for the few directions it was trained on. A solution to this is increasing the size of our training sets and diversifying the start and end locations. Another reason for these loops is that initial steps for the agent are harder to decide since it is further away from the goal. Because the percentage of loops in paths increased for our 10-hop datasets, this seems to be a plausible explanation. However, since the 10-hop datasets also have longer paths, there is more room for error and making wrong turns.

{\setlength\intextsep{0pt}
\begin{figure}[!t]
\includegraphics[height=2in]{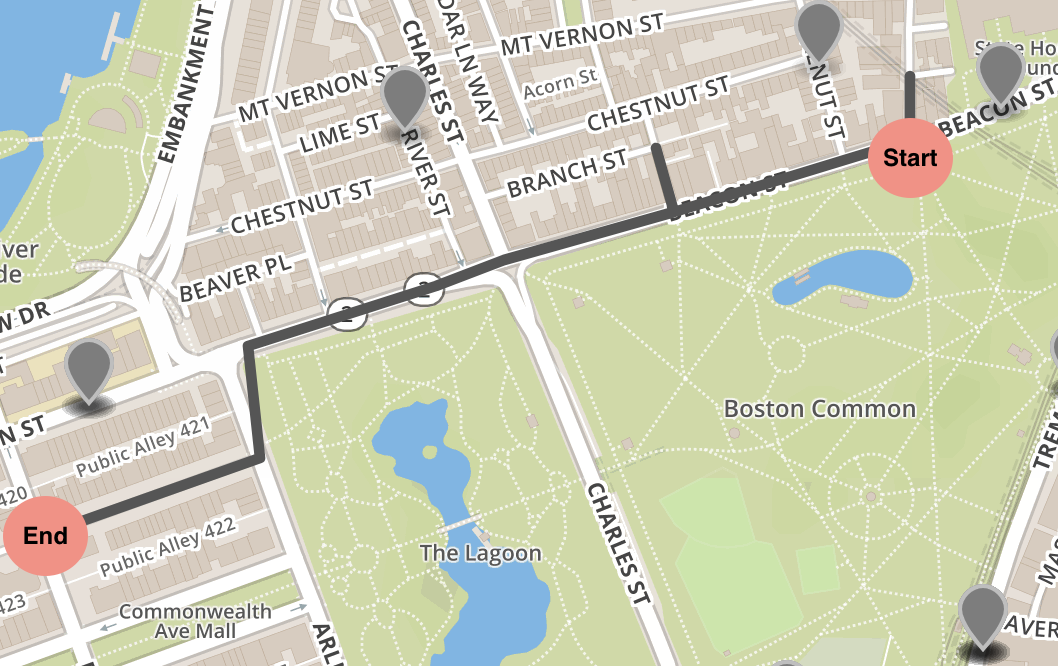}
\caption{Example of looping in a path generated by our model. The agent makes two wrong turns along the path in its initial steps.}
\label{fig:error}
\end{figure}
}%

Several extensions and studies are possible for SafeRoute. In the future, we plan to further optimize SafeRoute to enable longer paths and larger maps. In addition, we will study the portability of the model across urban environments. Currently, we evaluate all of our crime points equally. In the future, we plan on adding severity levels for our crimes to further develop the usefulness of our model. As discussed in Section~\ref{sec:relatedwork}, there has been much research in visual navigation. Therefore, another extension we plan for is to augment our agent with additional information to inform its decision, namely Google Street View images of the neighborhood. By combining multiple sources of information we hope to further boost the real-life relevance of our model. Another interesting direction would be to harness human feedback to the generated paths in our evaluation of the quality of the paths. We hope to not only validate the results of our work but also provide the generated feedback to an Inverse-Reinforcement learning framework to determine differences between our reward structure and the inferred one.

\section{Conclusion}\label{sec:conclusion}
In this project, we developed and tested a deep reinforcement learning model that aims to prevent street harassment by routing users away from local crime areas. We showed that an agent can learn to route users around high-density crime areas without strict supervision and can do so based on external data not contained within the graph. Our model can produce high-quality paths between two nodes in the graph that co-optimize to avoid crime areas without unreasonably sacrificing short travel distances and surpass our baselines in evaluation. With personal safety as a continual issue in people's lives, we hope SafeRoute can be a useful step in crime avoidance and helping users reach their destinations safely.

\bibliographystyle{ACM-Reference-Format}
\bibliography{main}


\begin{thebibliography}{00}


\ifx \showCODEN    \undefined \def \showCODEN     #1{\unskip}     \fi
\ifx \showDOI      \undefined \def \showDOI       #1{#1}\fi
\ifx \showISBNx    \undefined \def \showISBNx     #1{\unskip}     \fi
\ifx \showISBNxiii \undefined \def \showISBNxiii  #1{\unskip}     \fi
\ifx \showISSN     \undefined \def \showISSN      #1{\unskip}     \fi
\ifx \showLCCN     \undefined \def \showLCCN      #1{\unskip}     \fi
\ifx \shownote     \undefined \def \shownote      #1{#1}          \fi
\ifx \showarticletitle \undefined \def \showarticletitle #1{#1}   \fi
\ifx \showURL      \undefined \def \showURL       {\relax}        \fi
\providecommand\bibfield[2]{#2}
\providecommand\bibinfo[2]{#2}
\providecommand\natexlab[1]{#1}
\providecommand\showeprint[2][]{arXiv:#2}

\bibitem[\protect\citeauthoryear{Brunner, Richter, Wang, and
  Wattenhofer}{Brunner et~al\mbox{.}}{2017}]%
        {brunner2017teaching}
\bibfield{author}{\bibinfo{person}{Gino Brunner}, \bibinfo{person}{Oliver
  Richter}, \bibinfo{person}{Yuyi Wang}, {and} \bibinfo{person}{Roger
  Wattenhofer}.} \bibinfo{year}{2017}\natexlab{}.
\newblock \showarticletitle{Teaching a Machine to Read Maps with Deep
  Reinforcement Learning}.
\newblock \bibinfo{journal}{{\em AAAI 2018\/}} (\bibinfo{year}{2017}).
\newblock


\bibitem[\protect\citeauthoryear{Dijkstra}{Dijkstra}{1959}]%
        {dijkstra1959note}
\bibfield{author}{\bibinfo{person}{Edsger~W Dijkstra}.}
  \bibinfo{year}{1959}\natexlab{}.
\newblock \showarticletitle{A note on two problems in connexion with graphs}.
\newblock \bibinfo{journal}{{\em Numerische mathematik\/}} \bibinfo{volume}{1},
  \bibinfo{number}{1} (\bibinfo{year}{1959}), \bibinfo{pages}{269--271}.
\newblock


\bibitem[\protect\citeauthoryear{Galbrun, Pelechrinis, and Terzi}{Galbrun
  et~al\mbox{.}}{2016}]%
        {Galbrun:2016:UNB:2894858.2895106}
\bibfield{author}{\bibinfo{person}{Esther Galbrun},
  \bibinfo{person}{Konstantinos Pelechrinis}, {and} \bibinfo{person}{Evimaria
  Terzi}.} \bibinfo{year}{2016}\natexlab{}.
\newblock \showarticletitle{Urban Navigation Beyond Shortest Route}.
\newblock \bibinfo{journal}{{\em Inf. Syst.\/}} \bibinfo{volume}{57},
  \bibinfo{number}{C} (\bibinfo{date}{April} \bibinfo{year}{2016}),
  \bibinfo{pages}{160--171}.
\newblock
\showISSN{0306-4379}


\bibitem[\protect\citeauthoryear{Grillo, Paluch, and Livingston}{Grillo
  et~al\mbox{.}}{2014}]%
        {hollaback}
\bibfield{author}{\bibinfo{person}{Maria Grillo}, \bibinfo{person}{Rebecca
  Paluch}, {and} \bibinfo{person}{Beth Livingston}.}
  \bibinfo{year}{2014}\natexlab{}.
\newblock \bibinfo{title}{Cornell International Survey on Street Harassment}.
\newblock   (\bibinfo{year}{2014}).
\newblock


\bibitem[\protect\citeauthoryear{Grover and Leskovec}{Grover and
  Leskovec}{2016}]%
        {grover2016node2vec}
\bibfield{author}{\bibinfo{person}{Aditya Grover} {and} \bibinfo{person}{Jure
  Leskovec}.} \bibinfo{year}{2016}\natexlab{}.
\newblock \showarticletitle{node2vec: Scalable feature learning for networks}.
  In \bibinfo{booktitle}{{\em Proceedings of the 22nd ACM SIGKDD international
  conference on Knowledge discovery and data mining}}. ACM,
  \bibinfo{pages}{855--864}.
\newblock


\bibitem[\protect\citeauthoryear{Hendawi, Rustum, Oliver, Hazel, Teredesai, and
  Ali}{Hendawi et~al\mbox{.}}{2015}]%
        {hendawi2015multi}
\bibfield{author}{\bibinfo{person}{Abdeltawab~M Hendawi},
  \bibinfo{person}{Aqeel Rustum}, \bibinfo{person}{Dev Oliver},
  \bibinfo{person}{David Hazel}, \bibinfo{person}{Ankur Teredesai}, {and}
  \bibinfo{person}{Mohamed Ali}.} \bibinfo{year}{2015}\natexlab{}.
\newblock \showarticletitle{Multi-preference Time Dependent Routing}.
\newblock \bibinfo{journal}{{\em Technical Report UWT-CDS-TR-2015--03-01,
  Center for Data Science, Institute of Technology, University of Washington,
  Tacoma, Washington, USA\/}} (\bibinfo{year}{2015}).
\newblock


\bibitem[\protect\citeauthoryear{Hussein, Gaber, Elyan, and Jayne}{Hussein
  et~al\mbox{.}}{2017}]%
        {Hussein:2017:ILS:3071073.3054912}
\bibfield{author}{\bibinfo{person}{Ahmed Hussein},
  \bibinfo{person}{Mohamed~Medhat Gaber}, \bibinfo{person}{Eyad Elyan}, {and}
  \bibinfo{person}{Chrisina Jayne}.} \bibinfo{year}{2017}\natexlab{}.
\newblock \showarticletitle{Imitation Learning: A Survey of Learning Methods}.
\newblock \bibinfo{journal}{{\em ACM Comput. Surv.\/}} \bibinfo{volume}{50},
  \bibinfo{number}{2}, Article \bibinfo{articleno}{21} (\bibinfo{date}{April}
  \bibinfo{year}{2017}), \bibinfo{numpages}{35}~pages.
\newblock
\showISSN{0360-0300}


\bibitem[\protect\citeauthoryear{Johnson and Sanburn}{Johnson and
  Sanburn}{2017}]%
        {time_crime}
\bibfield{author}{\bibinfo{person}{David Johnson} {and} \bibinfo{person}{Josh
  Sanburn}.} \bibinfo{year}{2017}\natexlab{}.
\newblock \showarticletitle{Violent Crime Is On the Rise in U.S. Cities}.
\newblock \bibinfo{journal}{{\em Time\/}} (\bibinfo{year}{2017}).
\newblock


\bibitem[\protect\citeauthoryear{Kim, Cha, and Sandholm}{Kim
  et~al\mbox{.}}{2014}]%
        {kim2014socroutes}
\bibfield{author}{\bibinfo{person}{Jaewoo Kim}, \bibinfo{person}{Meeyoung Cha},
  {and} \bibinfo{person}{Thomas Sandholm}.} \bibinfo{year}{2014}\natexlab{}.
\newblock \showarticletitle{SocRoutes: safe routes based on tweet sentiments}.
  In \bibinfo{booktitle}{{\em Proceedings of the 23rd International Conference
  on World Wide Web}}. ACM, \bibinfo{pages}{179--182}.
\newblock


\bibitem[\protect\citeauthoryear{Kingma and Ba}{Kingma and Ba}{2014}]%
        {kingma2014adam}
\bibfield{author}{\bibinfo{person}{Diederik~P Kingma} {and}
  \bibinfo{person}{Jimmy Ba}.} \bibinfo{year}{2014}\natexlab{}.
\newblock \showarticletitle{Adam: A method for stochastic optimization}.
\newblock \bibinfo{journal}{{\em ICLR 2015\/}} (\bibinfo{year}{2014}).
\newblock


\bibitem[\protect\citeauthoryear{Kriegel, Renz, and Schubert}{Kriegel
  et~al\mbox{.}}{2010}]%
        {kriegel2010route}
\bibfield{author}{\bibinfo{person}{Hans-Peter Kriegel},
  \bibinfo{person}{Matthias Renz}, {and} \bibinfo{person}{Matthias Schubert}.}
  \bibinfo{year}{2010}\natexlab{}.
\newblock \showarticletitle{Route skyline queries: A multi-preference path
  planning approach}. In \bibinfo{booktitle}{{\em Data Engineering (ICDE), 2010
  IEEE 26th International Conference on}}. IEEE, \bibinfo{pages}{261--272}.
\newblock


\bibitem[\protect\citeauthoryear{Mata, Torres-Ruiz, and Guzman}{Mata
  et~al\mbox{.}}{2016}]%
        {mexico}
\bibfield{author}{\bibinfo{person}{Felix Mata}, \bibinfo{person}{Miguel
  Torres-Ruiz}, {and} \bibinfo{person}{Giovanni Guzman}.}
  \bibinfo{year}{2016}\natexlab{}.
\newblock \showarticletitle{A Mobile Information System Based on Crowd-Sensed
  and Official Crime Data for Finding Safe Routes: A Case Study of Mexico
  City}.
\newblock \bibinfo{journal}{{\em Mobile Information Systems\/}}
  \bibinfo{volume}{2016}, Article \bibinfo{articleno}{8068209}
  (\bibinfo{year}{2016}).
\newblock


\bibitem[\protect\citeauthoryear{Mirowski, Grimes, Malinowski, Hermann,
  Anderson, Teplyashin, Simonyan, Kavukcuoglu, Zisserman, and Hadsell}{Mirowski
  et~al\mbox{.}}{2018}]%
        {DBLP:journals/corr/abs-1804-00168}
\bibfield{author}{\bibinfo{person}{Piotr Mirowski},
  \bibinfo{person}{Matthew~Koichi Grimes}, \bibinfo{person}{Mateusz
  Malinowski}, \bibinfo{person}{Karl~Moritz Hermann}, \bibinfo{person}{Keith
  Anderson}, \bibinfo{person}{Denis Teplyashin}, \bibinfo{person}{Karen
  Simonyan}, \bibinfo{person}{Koray Kavukcuoglu}, \bibinfo{person}{Andrew
  Zisserman}, {and} \bibinfo{person}{Raia Hadsell}.}
  \bibinfo{year}{2018}\natexlab{}.
\newblock \showarticletitle{Learning to Navigate in Cities Without a Map}.
\newblock \bibinfo{journal}{{\em CoRR\/}}  \bibinfo{volume}{abs/1804.00168}
  (\bibinfo{year}{2018}).
\newblock
\showeprint[arxiv]{1804.00168}
\showURL{%
\url{http://arxiv.org/abs/1804.00168}}


\bibitem[\protect\citeauthoryear{{OpenStreetMap contributors}}{{OpenStreetMap
  contributors}}{2017}]%
        {OpenStreetMap}
\bibfield{author}{\bibinfo{person}{{OpenStreetMap contributors}}.}
  \bibinfo{year}{2017}\natexlab{}.
\newblock \bibinfo{title}{{Planet dump retrieved from https://planet.osm.org}}.
\newblock \bibinfo{howpublished}{\url{https://www.openstreetmap.org}}.
  (\bibinfo{year}{2017}).
\newblock


\bibitem[\protect\citeauthoryear{Quercia, Schifanella, and Aiello}{Quercia
  et~al\mbox{.}}{2014}]%
        {quercia2014shortest}
\bibfield{author}{\bibinfo{person}{Daniele Quercia}, \bibinfo{person}{Rossano
  Schifanella}, {and} \bibinfo{person}{Luca~Maria Aiello}.}
  \bibinfo{year}{2014}\natexlab{}.
\newblock \showarticletitle{The shortest path to happiness: Recommending
  beautiful, quiet, and happy routes in the city}. In \bibinfo{booktitle}{{\em
  Proceedings of the 25th ACM conference on Hypertext and social media}}. ACM,
  \bibinfo{pages}{116--125}.
\newblock


\bibitem[\protect\citeauthoryear{Rosenblatt}{Rosenblatt}{1956}]%
        {rosenblatt1956}
\bibfield{author}{\bibinfo{person}{Murray Rosenblatt}.}
  \bibinfo{year}{1956}\natexlab{}.
\newblock \showarticletitle{Remarks on Some Nonparametric Estimates of a
  Density Function}.
\newblock \bibinfo{journal}{{\em Ann. Math. Statist.\/}} \bibinfo{volume}{27},
  \bibinfo{number}{3} (\bibinfo{date}{09} \bibinfo{year}{1956}),
  \bibinfo{pages}{832--837}.
\newblock


\bibitem[\protect\citeauthoryear{Savinov, Dosovitskiy, and Koltun}{Savinov
  et~al\mbox{.}}{2018}]%
        {savinov2018semiparametric}
\bibfield{author}{\bibinfo{person}{Nikolay Savinov}, \bibinfo{person}{Alexey
  Dosovitskiy}, {and} \bibinfo{person}{Vladlen Koltun}.}
  \bibinfo{year}{2018}\natexlab{}.
\newblock \showarticletitle{Semi-parametric topological memory for navigation}.
  In \bibinfo{booktitle}{{\em International Conference on Learning
  Representations}}.
\newblock
\showURL{%
\url{https://openreview.net/forum?id=SygwwGbRW}}


\bibitem[\protect\citeauthoryear{Silver, Huang, Maddison, Guez, Sifre, van~den
  Driessche, Schrittwieser, Antonoglou, Panneershelvam, Lanctot, Dieleman,
  Grewe, Nham, Kalchbrenner, Sutskever, Lillicrap, Leach, Kavukcuoglu, Graepel,
  and Hassabis}{Silver et~al\mbox{.}}{2016}]%
        {Silver_2016}
\bibfield{author}{\bibinfo{person}{David Silver}, \bibinfo{person}{Aja Huang},
  \bibinfo{person}{Chris~J. Maddison}, \bibinfo{person}{Arthur Guez},
  \bibinfo{person}{Laurent Sifre}, \bibinfo{person}{George van~den Driessche},
  \bibinfo{person}{Julian Schrittwieser}, \bibinfo{person}{Ioannis Antonoglou},
  \bibinfo{person}{Veda Panneershelvam}, \bibinfo{person}{Marc Lanctot},
  \bibinfo{person}{Sander Dieleman}, \bibinfo{person}{Dominik Grewe},
  \bibinfo{person}{John Nham}, \bibinfo{person}{Nal Kalchbrenner},
  \bibinfo{person}{Ilya Sutskever}, \bibinfo{person}{Timothy Lillicrap},
  \bibinfo{person}{Madeleine Leach}, \bibinfo{person}{Koray Kavukcuoglu},
  \bibinfo{person}{Thore Graepel}, {and} \bibinfo{person}{Demis Hassabis}.}
  \bibinfo{year}{2016}\natexlab{}.
\newblock \showarticletitle{Mastering the Game of {Go} with Deep Neural
  Networks and Tree Search}.
\newblock \bibinfo{journal}{{\em Nature\/}} \bibinfo{volume}{529},
  \bibinfo{number}{7587} (\bibinfo{date}{Jan.} \bibinfo{year}{2016}),
  \bibinfo{pages}{484--489}.
\newblock


\bibitem[\protect\citeauthoryear{SpotCrime}{SpotCrime}{2007}]%
        {spotcrime}
SpotCrime \bibinfo{year}{2007}\natexlab{}.
\newblock \bibinfo{howpublished}{\url{https://spotcrime.com/}}.
  (\bibinfo{year}{2007}).
\newblock


\bibitem[\protect\citeauthoryear{Wang, Xiong, Wang, and Wang}{Wang
  et~al\mbox{.}}{2018}]%
        {DBLP:journals/corr/abs-1803-07729}
\bibfield{author}{\bibinfo{person}{Xin Wang}, \bibinfo{person}{Wenhan Xiong},
  \bibinfo{person}{Hongmin Wang}, {and} \bibinfo{person}{William~Yang Wang}.}
  \bibinfo{year}{2018}\natexlab{}.
\newblock \showarticletitle{Look Before You Leap: Bridging Model-Free and
  Model-Based Reinforcement Learning for Planned-Ahead Vision-and-Language
  Navigation}.
\newblock \bibinfo{journal}{{\em CoRR\/}}  \bibinfo{volume}{abs/1803.07729}
  (\bibinfo{year}{2018}).
\newblock
\showeprint[arxiv]{1803.07729}
\showURL{%
\url{http://arxiv.org/abs/1803.07729}}


\bibitem[\protect\citeauthoryear{Wei, Ma, and Jia}{Wei et~al\mbox{.}}{2014}]%
        {wei2014day}
\bibfield{author}{\bibinfo{person}{Fangfang Wei}, \bibinfo{person}{Shoufeng
  Ma}, {and} \bibinfo{person}{Ning Jia}.} \bibinfo{year}{2014}\natexlab{}.
\newblock \showarticletitle{A day-to-day route choice model based on
  reinforcement learning}.
\newblock \bibinfo{journal}{{\em Mathematical Problems in Engineering\/}}
  \bibinfo{volume}{2014} (\bibinfo{year}{2014}).
\newblock


\bibitem[\protect\citeauthoryear{Williams}{Williams}{1992}]%
        {Williams1992}
\bibfield{author}{\bibinfo{person}{Ronald~J. Williams}.}
  \bibinfo{year}{1992}\natexlab{}.
\newblock \showarticletitle{Simple statistical gradient-following algorithms
  for connectionist reinforcement learning}.
\newblock \bibinfo{journal}{{\em Machine Learning\/}} \bibinfo{volume}{8},
  \bibinfo{number}{3} (\bibinfo{date}{01 May} \bibinfo{year}{1992}),
  \bibinfo{pages}{229--256}.
\newblock
\showISSN{1573-0565}
\showDOI{%
\url{https://doi.org/10.1007/BF00992696}}


\bibitem[\protect\citeauthoryear{Xiong, Hoang, and Wang}{Xiong
  et~al\mbox{.}}{2017}]%
        {wenhan_emnlp2017}
\bibfield{author}{\bibinfo{person}{Wenhan Xiong}, \bibinfo{person}{Thien
  Hoang}, {and} \bibinfo{person}{William~Yang Wang}.}
  \bibinfo{year}{2017}\natexlab{}.
\newblock \showarticletitle{DeepPath: A Reinforcement Learning Method for
  Knowledge Graph Reasoning}. In \bibinfo{booktitle}{{\em Proceedings of the
  2017 Conference on Empirical Methods in Natural Language Processing (EMNLP
  2017)}}. \bibinfo{publisher}{ACL}, \bibinfo{address}{Copenhagen, Denmark}.
\newblock


\bibitem[\protect\citeauthoryear{Yang and Diez-Roux}{Yang and
  Diez-Roux}{2012}]%
        {yang2012walking}
\bibfield{author}{\bibinfo{person}{Yong Yang} {and} \bibinfo{person}{Ana~V
  Diez-Roux}.} \bibinfo{year}{2012}\natexlab{}.
\newblock \showarticletitle{Walking distance by trip purpose and population
  subgroups}.
\newblock \bibinfo{journal}{{\em American journal of preventive medicine\/}}
  \bibinfo{volume}{43}, \bibinfo{number}{1} (\bibinfo{year}{2012}),
  \bibinfo{pages}{11--19}.
\newblock


\bibitem[\protect\citeauthoryear{Yuan, Zheng, Zhang, Xie, Xie, Sun, and
  Huang}{Yuan et~al\mbox{.}}{2010}]%
        {Yuan:2010:TDD:1869790.1869807}
\bibfield{author}{\bibinfo{person}{Jing Yuan}, \bibinfo{person}{Yu Zheng},
  \bibinfo{person}{Chengyang Zhang}, \bibinfo{person}{Wenlei Xie},
  \bibinfo{person}{Xing Xie}, \bibinfo{person}{Guangzhong Sun}, {and}
  \bibinfo{person}{Yan Huang}.} \bibinfo{year}{2010}\natexlab{}.
\newblock \showarticletitle{T-drive: Driving Directions Based on Taxi
  Trajectories}. In \bibinfo{booktitle}{{\em Proceedings of the 18th SIGSPATIAL
  International Conference on Advances in Geographic Information Systems}} {\em
  (\bibinfo{series}{GIS '10})}. \bibinfo{publisher}{ACM}, \bibinfo{address}{New
  York, NY, USA}, \bibinfo{pages}{99--108}.
\newblock
\showISBNx{978-1-4503-0428-3}


\bibitem[\protect\citeauthoryear{Zaremba and Sutskever}{Zaremba and
  Sutskever}{2015}]%
        {DBLP:journals/corr/ZarembaS15}
\bibfield{author}{\bibinfo{person}{Wojciech Zaremba} {and}
  \bibinfo{person}{Ilya Sutskever}.} \bibinfo{year}{2015}\natexlab{}.
\newblock \showarticletitle{Reinforcement Learning Neural Turing Machines}.
\newblock \bibinfo{journal}{{\em CoRR\/}}  \bibinfo{volume}{abs/1505.00521}
  (\bibinfo{year}{2015}).
\newblock
\showeprint[arxiv]{1505.00521}
\showURL{%
\url{http://arxiv.org/abs/1505.00521}}


\end{thebibliography}

\end{document}